\documentclass[letterpaper, 10 pt, conference]{ieeeconf}

\IEEEoverridecommandlockouts                              

\usepackage{cite}
\usepackage{tabularx}
\usepackage{colortbl}
\usepackage{multirow}
\usepackage{multirow}
\usepackage{amsmath,amssymb,amsfonts}
\usepackage{mathtools, bigstrut}
\usepackage{graphicx}
\usepackage{textcomp}
\usepackage{wrapfig}

\usepackage[linesnumbered,ruled,lined]{algorithm2e}

\usepackage{xcolor}
\usepackage{soul}

\usepackage{multirow} 
\usepackage{makecell} 
\usepackage{booktabs} 
\usepackage{array} %
\usepackage{colortbl}

\usepackage{authblk}
\let\labelindent\relax
\usepackage{enumitem}
\usepackage[pagebackref]{hyperref}
\hypersetup{
    colorlinks=true,
    citecolor=ForestGreen,
    linkcolor=blue,
    filecolor=magenta,      
    urlcolor=cyan
}
\usepackage{breqn}
\usepackage{algpseudocode}
\usepackage{caption}
\usepackage{subcaption}
\usepackage{tcolorbox}
\usepackage{array}
\usepackage{makecell}
\usepackage{cuted}
\usepackage{epsfig}

\usepackage[english]{babel}
\usepackage{amsthm}

\usepackage{tabularray}
\definecolor{Silver}{rgb}{0.752,0.752,0.752}
\definecolor{Gallery}{rgb}{0.937,0.937,0.937}
\usepackage{adjustbox}

\allowdisplaybreaks

\title{\LARGE \bf CrowdSurfer: Sampling Optimization Augmented with Vector-Quantized Variational AutoEncoder for Dense Crowd Navigation}

\author{Naman Kumar*$^{1}$, Antareep Singha*$^{1}$, Laksh Nanwani*$^{1}$, Dhruv Potdar$^{1}$, Tarun R$^{1}$, Fatemeh Rastgar$^{2}$,\\Simon Idoko$^{3}$, Arun Kumar Singh$^{3}$, K. Madhava Krishna$^{1}$
\thanks{* denotes equal contribution}
\thanks{$^{1}$ are with Robotics Research Center, IIIT Hyderabad, India.}
\thanks{$^{2}$ is with the Department of Natural Sciences and Technology, Orebro University, Sweden}
\thanks{$^{3}$ are with University of Tartu, Estonia}
\thanks{
Emails: (namanxkumar, antareepsinha12, lakshanshul, dhruvpotdar29, tarun.ramak)@gmail.com, fatemeh.rastgar@oru.se, simon.idoko@ut.ee, 
aks1812@gmail.com, mkrishna@iiit.ac.in.}
\thanks{Project Page-\href{https://smart-wheelchair-rrc.github.io/CrowdSurfer-webpage}{https://smart-wheelchair-rrc.github.io/CrowdSurfer-webpage}}
\thanks{Code-\href{https://github.com/Smart-Wheelchair-RRC/CrowdSurfer}{https://github.com/Smart-Wheelchair-RRC/CrowdSurfer}}
}

\makeatletter

\date{September 2024}

\begin{document}



\maketitle
\begin{strip}
\begin{minipage}{\textwidth}\centering
\vspace{-70pt}
\includegraphics[width=\linewidth]{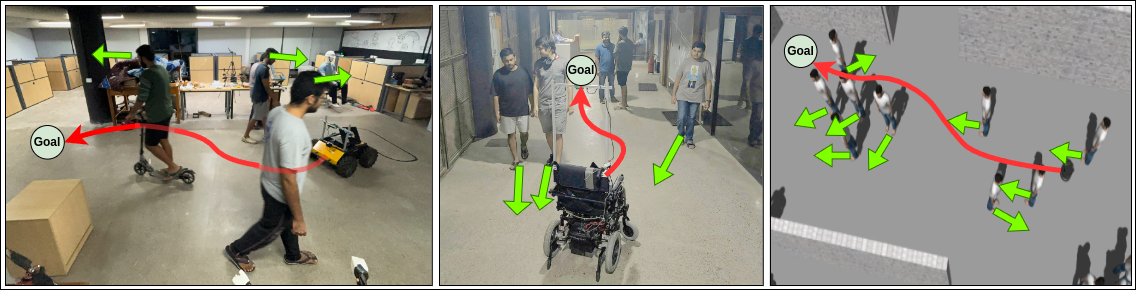}
\captionof{figure}{The proposed method, \textbf{CrowdSurfer}, first generates a diverse set of samples using the VQ-VAE + PixelCNN pipeline, followed by inference-time optimization via the PRIEST planner \cite{rastgar2024priest}, resulting in trajectories that can traverse complex dynamic navigation scenarios successfully. From left to right, we show HuskyA200, our Autonomous Wheelchair, and a simulated turtlebot navigating in crowded environments using \textbf{CrowdSurfer}.}
\label{fig:fig_teaser}
\end{minipage}
\end{strip}

\thispagestyle{empty}
\pagestyle{empty}


\begin{abstract}
Navigation amongst densely packed crowds remains a challenge for mobile robots. The complexity increases further if the environment layout changes, making the prior computed global plan infeasible. In this paper, we show that it is possible to dramatically enhance crowd navigation by just improving the local planner. Our approach combines generative modelling with inference-time optimization to generate sophisticated long-horizon local plans at interactive rates. More specifically, we 
train a Vector Quantized Variational AutoEncoder to learn a prior over the expert trajectory distribution conditioned on the perception input. At run-time, this is used as an initialization for a sampling-based optimizer for further refinement. Our approach does not require any sophisticated prediction of dynamic obstacles and yet provides state-of-the-art performance. In particular, we compare against the recent DRL-VO approach \cite{xie2023drl} and show a $40\%$ improvement in success rate and a $6\%$ improvement in travel time.
   
\end{abstract}

\section{Introduction}
Reliable, collision-free navigation in complex environments filled with human crowds is essential for deploying mobile robots in hospitals, offices, airports, etc \cite{valner2022scalable}, \cite{pratkanis2013replacing}. While this is a well-studied problem, one understated aspect of existing works is the reliance on a global plan computed based on prior maps. Put differently, existing planners show poor performance in the absence of a global plan or when it is rendered infeasible due to changes in the environment \cite{rastgar2024priest}, \cite{xiao2022autonomous}. 

The presence of dense human crowds further necessitates complex on-the-fly decision-making, as prior plans are of little use in this context. Existing works have tried to heavily leverage both imitation \cite{bi2018navigation} and reinforcement learning (RL) \cite{chen2019crowd}, \cite{zhou2022robot}, \cite{chen2020relational} for navigation amongst dense crowds. The latter is particularly attractive since it not only gets rid of the necessity of obtaining expert demonstration but can also implicitly account for the interaction between the robot and the crowd. However, it is worth pointing out that few works like \cite{xie2023drl} have shown navigation amongst dense crowds in indoor environments with very tight spaces.

This paper shows that we can dramatically improve navigation in crowded tight spaces by just making the local planner more capable. In particular, a local planner capable of long-horizon planning at interactive rates can deliver strong performance even without any complex trajectory or interaction prediction of human crowds. To this end, our key \textbf{contribution} in this paper is the development of a local planner that combines generative modelling with inference-time optimization. Specifically, we use Vector Quantized Variational AutoEncoder (VQ-VAE) \cite{van2017neural},  to learn a discrete prior over the space of expert demonstrations. The discrete latent space is particularly suitable for capturing the inherent multi-modality of the expert demonstrations. We also train a PixelCNN \cite{van2016conditional} to sample from the learned priors and generate a distribution of trajectories during inference, conditioned on the perception input (the local occupancy map, pedestrian positions and velocities, and the heading to the goal). 

We also perform inference-time optimizations to ensure that the PixelCNN-generated trajectories exactly satisfy kinematic and collision constraints. This is achieved by using the learned posterior distribution as an initialization for a sampling-based trajectory optimizer built on top of our prior work \cite{rastgar2024priest}. A unique aspect of \cite{rastgar2024priest} is that it combines convex optimization and gradient-free search to search different potential homotopies for collision avoidance.

Although conceptually simple, our approach performs strongly in various open-source benchmark environments augmented with human crowds controlled by the social-force model \cite{shafabakhsh2013simulation}. We specifically compare against DRL-VO \cite{xie2023drl} that combines RL with velocity-obstacle \cite{van2008reciprocal} based reactive collision avoidance. We achieve $40\%$ and $6\%$  improvement over DRL-VO in success rate and travel time, respectively. 

\section{Related Works}
Autonomous navigation capabilities still lack repeatability and robustness, especially in cluttered environments with dense crowds. Nevertheless, this problem has been well studied, and we have reviewed the prominent works in this direction and contrasted our approach with them.


\noindent \textbf{Model-Based Planning:} This class of planners is based on classical graph/sampling-based search, mathematical optimization, or a combination of both. Two such approaches that still form the backbone of autonomous navigation in both research and industry are Dynamic Window Approach (DWA) \cite{fox1997dynamic} and Timed Elastic Bands (TEB) \cite{rosmann2015timed}. In our recent paper \cite{rastgar2024priest}, we showed that these two approaches substantially outperform even some recent approaches like Model Predictive Path Integral \cite{williams2017model} and its variants like Log-MPPI \cite{mohamed2022autonomous}. A similar observation was also put forward in \cite{xiao2022autonomous}. An important caveat in model-based planning is the requirement of a global planner, which in turn requires a prior map of the environments. In the absence of a global plan, all existing approaches show massive performance deterioration, even without dynamic obstacles \cite{rastgar2024priest}, \cite{xiao2022autonomous}.

\noindent \textbf{Learning-Based Approaches:} Over the last decade, there has been a strong interest in applying RL to crowd navigation  \cite{chen2019crowd}, \cite{zhou2022robot}, \cite{chen2020relational}, \cite{everett2018motion}, \cite{9197379}, \cite{9099106}. These works rely on implicitly or explicitly modelling the interaction between the crowd and the robot and that between members of the crowd themselves. However, most RL approaches show results in spaces where the robot just needs to deal with the dynamic crowd. Recently, \cite{xie2023drl} has shown impressive performance in environments that resemble real-world settings that are highly cluttered as well as have tens of dynamic human obstacles. One critical drawback of purely learning-based approaches is that they typically struggle when encountered with novel scenarios. As we show later, even the performance of \cite{xie2023drl} deteriorates when tasked with longer runs.

Social navigation frameworks like \cite{hirose2023sacson}, \cite{8011466}, \cite{6698863}, \cite{10611710}, \cite{10610025} focus on collision avoidance while also adhering to social norms, enabling robots to anticipate and respect human movement/interaction in social spaces. In contrast, the focus of our work is to address the challenges of local navigation, which in turn, can also contribute towards improving human-aware navigation.

\noindent \textbf{Our Contribution Over SOTA:} Our proposed approach aims to find the middle ground between model-based planning and purely learning-based approaches. We aim to make the algorithmic parameters of the model-based planners adaptive to the environmental conditions. At the same time, we intend to improve performance in novel scenarios. We fulfil both these objectives by combining generative model (VQ-VAE+PixelCNN) based imitation learning with an expressive model-based planner PRIEST \cite{rastgar2024priest} capable of searching across multiple homotopies. As mentioned earlier, the VQ-VAE+PixelCNN can be seen as learning a prior over the optimal trajectories while PRIEST provides the inference-time refinement.

\section{Methodology}

\subsection{Problem Formulation}
\noindent We follow the trajectory optimization template and formulate crowd navigation in the following manner.

\begin{figure*}[h!]
    \centering
    \includegraphics[width=0.8\linewidth]{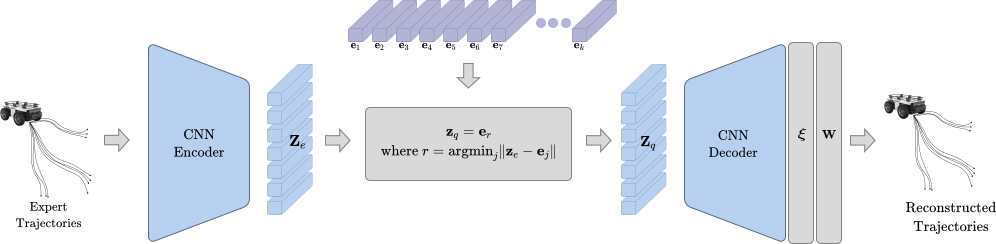}
    \caption{\footnotesize{The VQ-VAE pipeline is used to learn the discrete latent space of optimal demonstration trajectories. We train the VQ-VAE to predict polynomial coefficients that are converted to trajectories through \eqref{parametrized}. This ensures that the reconstructed trajectories have higher continuity and differentiability. }}
    \label{fig_vq}
    \vspace{-0.5cm}
\end{figure*}

\vspace{-0.45cm}
\small
\begin{subequations}
\begin{align}
   \hspace{-2.5cm}\min_{x(t), y(t)} c(x^{(q)}(t), y^{(q)}(t)),  
\label{acc_cost}\\
   x^{(\hspace{-0.02cm}q\hspace{-0.02cm})}\hspace{-0.05cm}(\hspace{-0.02cm}t_0\hspace{-0.02cm}), \hspace{-0.05cm}y^{(\hspace{-0.02cm}q\hspace{-0.02cm})}\hspace{-0.05cm}(\hspace{-0.02cm}t_0\hspace{-0.02cm})=\hspace{-0.05cm}\textbf{b}_{0},~
   x^{(\hspace{-0.02cm}q\hspace{-0.02cm})}\hspace{-0.05cm}(\hspace{-0.02cm}t_f\hspace{-0.02cm}),\hspace{-0.05cm}y^{(\hspace{-0.02cm}q\hspace{-0.02cm})}\hspace{-0.05cm}(\hspace{-0.02cm}t_f\hspace{-0.02cm}) = \hspace{-0.05cm}\textbf{b}_{f},\label{eq1_multiagent_1}\\
    \dot{x}^{2}(\hspace{-0.02cm}t\hspace{-0.02cm})\hspace{-0.05cm}+\hspace{-0.05cm} \dot{y}^{2}(\hspace{-0.02cm}t\hspace{-0.02cm})\hspace{-0.05cm}\leq v^{2}_{max}, ~
    \ddot{x}^{2}(\hspace{-0.02cm}t\hspace{-0.02cm}) \hspace{-0.05cm}+\hspace{-0.05cm} \ddot{y}^{2}(\hspace{-0.02cm}t\hspace{-0.02cm}) \hspace{-0.05cm}\leq a^{2}_{max}, \label{acc_constraint}\\
    \hspace{-0.4cm}
    -\frac{\hspace{-0.07cm}(x(t)\hspace{-0.09cm}-
 \hspace{-0.06cm}x_{o, j}(t)\hspace{-0.01cm})^{2}}{a_j^2}\hspace{-0.09cm} - \hspace{-0.09cm}
 \frac{\hspace{-0.07cm}(y(t) \hspace{-0.075cm}-\hspace{-0.065cm}y_{o, j}(t)\hspace{-0.02cm})^{2}}{a_j^2} 
  \hspace{-0.09cm}
  + \hspace{-0.05cm}1\leq 0, \label{coll_multiagent}
\end{align}
\end{subequations}
\normalsize

\noindent where $(x(t),y(t))$ and $(x_{o, j}(t),y_{o, j}(t))$~respectively denote the robot and the $j^{th}$ obstacle position at time $t$. These obstacles could be either individual LiDAR points or a dynamic human. We model each type as a circular obstacle with radius $a_j$. The function $c(.)$ could be any arbitrary function (even non-smooth, non-analytic)
of derivatives of the position-level trajectories. We can also leverage the differential flatness of a typical mobile robot to include control costs in $c(.)$. The vectors $\textbf{b}_{0}$ and $\textbf{b}_{f}$ in \eqref{eq1_multiagent_1} represent the initial and final values of boundary condition on the $q^{th}$ derivative of the~position-level trajectory. We employ $q\hspace{-0.10cm}=\hspace{-0.10cm}\{0,1,2\}$ in our implementation. Inequalities \eqref{acc_constraint} bound the maximum values of velocities and accelerations. In \eqref{coll_multiagent}, we enforce collision avoidance, assuming circular obstacle with radius $a_j$.

We can convert \eqref{acc_cost}-\eqref{coll_multiagent} into a finite-dimensional representation by parameterizing the trajectories as polynomials in the following manner.

\vspace{-0.25cm}

\begin{align}
    \begin{bmatrix}
    x(t_{1}) \\ \vdots \\ x(t_{n_{p}})
    \end{bmatrix}^{T} \hspace{-0.3cm} = \mathbf{P} \hspace{0.1cm}\bold{c}_{x} ,\hspace{-0.1cm}
     \begin{bmatrix}
    y(t_{1})\\ \vdots\\ y(t_{n_{p}})
    \end{bmatrix}^{T} \hspace{-0.2cm}= \mathbf{P} \bold{c}_{y}, \mathbf{W} = \begin{bmatrix}
        \mathbf{P} & \mathbf{0}\\
        \mathbf{0} & \mathbf{P}
    \end{bmatrix}\label{parametrized}
\end{align}

\noindent where the matrix $\mathbf{P}$ is a matrix formed by time-dependent polynomial basis functions. The vectors $\mathbf{c}_x, \mathbf{c}_y$ are the coefficients attached to the individual basis functions. We can represent the derivatives like $\dot{x}(t), \ddot{x}(t), \dot{y}(t), \ddot{y}(t) $ in a similar manner as \eqref{parametrized} using $\dot{\mathbf{P}}$ and $\ddot{\mathbf{P}}$

Using \eqref{parametrized}, we can represent \eqref{acc_cost}-\eqref{coll_multiagent} in the following compact form, wherein $\boldsymbol{\xi} = (\mathbf{c}_x, \mathbf{c}_y)$

\vspace{-0.55cm}
\begin{subequations}
\begin{align}   \min_{\hspace{0.1cm}\boldsymbol{\xi}} c(\boldsymbol{\xi})  \label{cost_modify} \\
    \textbf{A}\boldsymbol{\xi} = \textbf{b}_{eq}\label{eq_modify}\\
   \textbf{g}(\boldsymbol{\xi}) \leq \textbf{0} \label{reform_bound},
   \vspace{-0.2cm}
\end{align}
\end{subequations}
\vspace{-0.5cm}



\subsection{Learning the Discrete Latent Space of Solutions Through VQ-VAE}
\noindent We intend to learn the solution space of \eqref{acc_cost}-\eqref{coll_multiagent} through demonstration of optimal trajectories in a similar style as our prior work \cite{idoko2024learning} designed for autonomous driving. To this end, we first compress the demonstration trajectories to a discrete latent space. Such representation can capture the multi-modality in the solution trajectories, such as avoiding obstacles from the left vis-a-vis from the right.

Our VQ-VAE architecture, illustrated in Fig. \ref{fig_vq} adapts the original formulation \cite{oord2017neural}  for trajectory generation. We use a CNN encoder to compress expert trajectories into a continuous latent space $\mathbf{Z}_e \in \mathbb{R}^{L \times D}$, where $L$ denotes the number of latent vectors, each with dimension $D$. We denote the $i^{th}$ latent vector in $\mathbf{Z}_e$ as $\mathbf{z}_{e, i}$. The fundamental feature of VQ-VAE is the transformation from continuous space to a discrete one: mapping $\mathbf{Z}_e$ to $\mathbf{Z}_q$. To achieve this, we introduce a latent embedding matrix $\mathbf{E} \in \mathbb{R}^{K \times D}$ called code-book with $K$ discrete vectors $\mathbf{e}_j$. Each $i^{th}$ latent vector in $\mathbf{Z}_e$ is assigned the nearest $\mathbf{e}_j$ vector based on the nearest neighbour equation detailed in \eqref{eqn_nearest}, which makes up the $i^{th}$ latent vector in $\mathbf{Z}_q$.

\vspace{-0.45cm}
\begin{multline}
    \boldsymbol{z}_{q,i} = \boldsymbol{e}_r, \text{ where } r = \arg \min_j || \boldsymbol{z}_{e, i} - \boldsymbol{e}_j ||_2^2   \\
    \text{for } i \in \{1, 2, ..., L\} 
    \label{eqn_nearest}
\end{multline} 

The VQ-VAE decoder receives $\textbf{Z}_q$ and reconstructs it to polynomial coefficients $\boldsymbol{\xi}$ and subsequently to trajectories using \eqref{parametrized}. The VQ-VAE model is optimized using the loss function detailed in \eqref{eqn_vq_loss}, which comprises three parts. The first term is the reconstruction loss which ensures that the encoder/decoder pair effectively reconstructs the input expert trajectory. The second and third terms are called the codebook and commitment loss, which are used to update the code-book vectors during training and bypass the non-differentiable discretization presented in \eqref{eqn_nearest}.


\vspace{-0.45cm}
\begin{multline}
    \mathcal{L}_{vqvae} = \| \mathbf{W} \boldsymbol{\xi} - \boldsymbol{\tau}_e \|_2^2 \;+\; \| \text{sg}[\mathbf{Z}_e] - \mathbf{E}\|_2^2 \\
    \;+\;  \beta \| \mathbf{Z}_e(x) - \text{sg}[\mathbf{E}]\|_2^2
    \label{eqn_vq_loss}
\end{multline} 

\vspace{0.1cm}
\subsection{Using a Conditional PixelCNN for sampling from the VQ-VAE latent space}

\begin{figure}[ht]
    \centering
    \includegraphics[width = 0.87\linewidth]{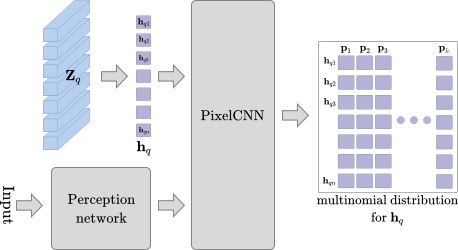}
    \caption{\footnotesize{It is possible to express the learned discrete latent space $\mathbf{Z}_q$ as a vector $\mathbf{h}_{q}$ whose elements signify which code-book vector is used to form the $i^{th}$ row of $\mathbf{Z}_q$. Our PixelCNN model outputs a multinomial probability distribution over $\mathbf{h}_{q}$. The number of discrete probabilities $p_k$ is equal to the number of code-book vectors. Thus, the PixelCNN model decides the probability that the $i^{th}$ element of $\mathbf{h}_q$ is formed by the $r^{th}$ code-book vector. Sampling from the multinomial distribution allows the generation of different $\mathbf{h}_q$, each leading to a distinct trajectory.  }  }
    \label{fig_px}
    \vspace{-0.2cm}
\end{figure}


\noindent For sampling from the latent space of the VQ-VAE, we adapt the Conditional PixelCNN \cite{van2016conditional} that allows us to model the conditional distributions of the space. Using this model, we can generate diverse trajectories (via the VQ-VAE decoder) for different environments in which the robot operates, by simply conditioning it on features extracted from the occupancy grid map, dynamic obstacle states, and immediate heading to the goal.

We recall that the discrete latent space is a matrix $\mathbf{Z}_q$. Each row of this matrix is formed by $\mathbf{z}_{q, i}$, which in turn is related to the $r^{th}$ code book vector through \eqref{eqn_nearest}. Thus, the information about $\mathbf{Z}_q$ is encoded into a vector $\mathbf{h}_q$ in which each element stores the index of the corresponding codebook vector used to form $\mathbf{z}_{q, i}$ (see Fig.\ref{fig_px}). Hence, the training of the VQ-VAE in the last section provides us with the ground-truth values of  $\mathbf{h}_q$.

The  PixelCNN model outputs a multinomial probability distribution over $\mathbf{h}_q$. We can sample from this distribution to generate different $\mathbf{h}_q$ samples and consequently diverse $\mathbf{Z}_{q}$. This, in turn, can be fed to the trained decoder of VQ-VAE to generate trajectory samples.

The defining feature of the PixelCNN model is that the probability distribution over $\mathbf{h}_q$ is generated in an auto-regressive manner through \eqref{auto_regressive}. That is, the prediction of each element of $\mathbf{h}_q$ depends on the prediction of the previous elements and the conditioning input. During the training phase, we form a cross-entropy loss over the ground-truth $\mathbf{h}_q$ (from VQ-VAE), and that is predicted by PixelCNN to learn the parameters of the multinomial distribution.
\vspace{-0.3mm}
\begin{align}
    p(\mathbf{h}_q|\mathcal{O}) = \prod_{i=1}^{L}p(\mathbf{h}_{q,i}|\mathbf{h}_{q,1},\hdots, \mathbf{h}_{q,i-1}, \mathcal{O})
    \label{auto_regressive}
\end{align}

In the inferencing phase, we start with $\mathbf{h}_q$ set to zeros and using the trained PixelCNN, we recursively generate indices $\mathbf{h}_{q,i}$ through \eqref{auto_regressive} based on the observed conditioning input, to obtain a multinomial distribution for each element of $\mathbf{h}_q$. 


\subsection{Inference-Time Optimization}
\begin{figure}[t!]
    \centering
    \includegraphics[width=\linewidth]{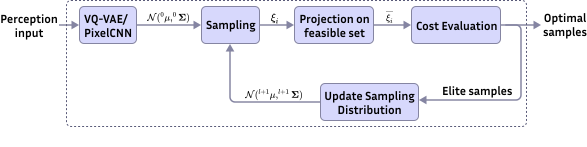}
    \caption{\footnotesize{ At any given planning step, the VQ-VAE+PixelCNN combination takes in the perception input and generates samples for initialization of the PRIEST planner \cite{rastgar2024priest}. The planner subsequently refines the predicted distribution using a combination of projection-optimizer augmented cross-entropy method \cite{wen2018constrained}  } }
    \label{fig_integration}
    \vspace{-0.4cm}
\end{figure}

\noindent Although VQ-VAE and PixelCNN are powerful generative models, their generated trajectories may not satisfy the different kinematic and collision constraints. Thus, we refine the PixelCNN-generated trajectory distribution with a sampling-based optimization PRIEST \cite{rastgar2024priest}. Specifically, we replace the Gaussian distribution initialization of PRIEST with our PixelCNN model (see Fig. \ref{fig_integration}). 

\subsection{Data Collection Pipeline and Perception Network Architecture}
\subsubsection{Data Collection Pipeline}
The training data for the pipeline includes expert trajectories of a robot navigating highly dynamic environments, used to train the unconditional VQ-VAE, and observation data for training the PixelCNN model. All data was collected in a simulated setting using the PEDSIM social-force library \cite{gloor-2016} and the Gazebo simulator \cite{koenig2004design}. We teleoperated a Turtlebot2 within the open-source Lobby World environment, featuring 35 dynamic agents. The simulated positions of these agents (agent states) and LiDAR scans were used as ground truth inputs. The teleoperated expert trajectories were pre-processed using the PRIEST planner. This resulted in smoother trajectories that proved more conducive to training VQ-VAE.

\begin{figure}[!h]
    \centering
    \includegraphics[width=\linewidth]{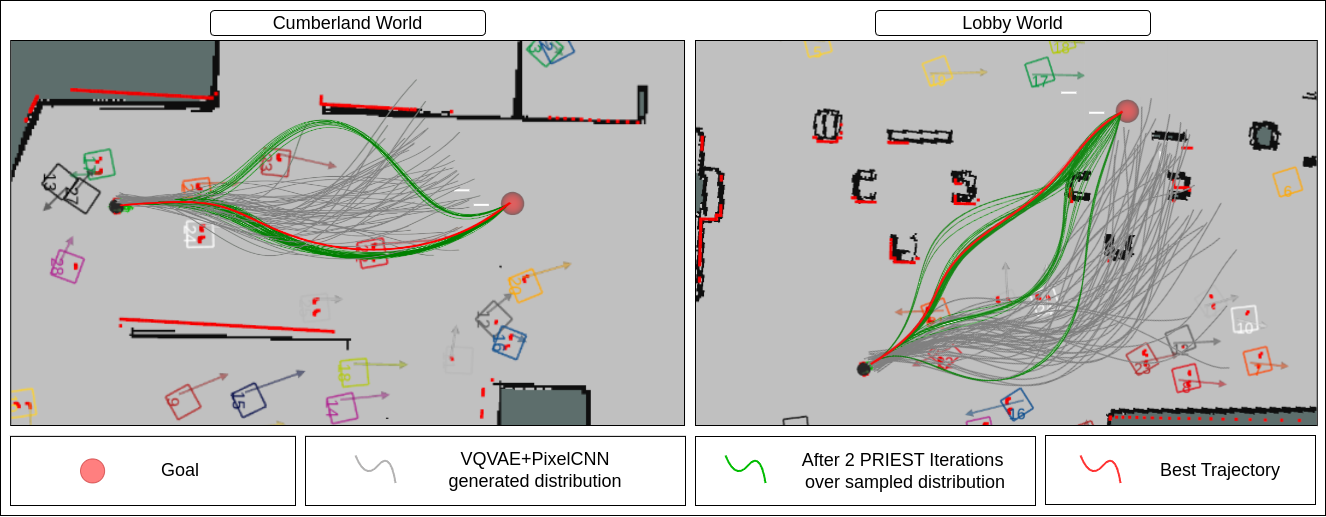}
    \caption{\footnotesize{Trajectories at 3 stages of the pipeline. VQVAE + PixelCNN generate diverse primitives (\textcolor{gray}{GREY}), that may not satisfy constraints or reach the goal. These trajectories are then optimized via 2 PRIEST \cite{rastgar2024priest} iterations (\textcolor{green}{GREEN}) to satisfy constraints, and the best trajectory as scored by PRIEST is chosen (\textcolor{red}{RED}).}}
    \label{fig_qualitative}
\end{figure}

\begin{figure}[ht]
    \centering
    \includegraphics[width=0.55\linewidth]
    {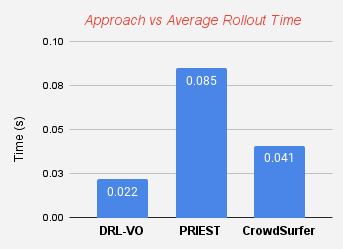}
    \caption{\footnotesize{CrowdSurfer improves upon the computation times of PRIEST while also delivering better results in navigation than both PRIEST (Table \ref{PRIEST Ablation}) and DRL-VO (Table \ref{Table1}). Our pipeline, being a trajectory rollout approach, still achieves real-time frequencies of approx. 20 Hz, making it easily deployable on real-world robots (Fig. \ref{fig:fig_teaser})}}
    \label{fig_graph}
    \vspace{-0.6cm}
\end{figure}

For PixelCNN training, the observation data comprising of LiDAR scans and dynamic agent states were preprocessed to match the network's input requirements. LiDAR scans were converted to occupancy maps and dynamic agent states were used to calculate agent velocities. Additionally, odometry data was processed to generate ego velocities and heading angle-to-goal.

\vspace{0.1mm}
\subsubsection{Perception Network Architecture}
The Perception Network encodes the observation space into a single conditioning vector which is passed to the PixelCNN. It accepts an occupancy map in the form of a single channel image $O \in \mathbb{R}^{H\times W\times 1}$, positions and velocities of dynamic obstacles for the past $T (=5)$ timesteps $D \in \mathbb{R}^{T\times 4\times M}$ where $M (=10)$ is the maximum number of dynamic obstacles. We also use the current heading of the mobile robot to the current goal in radians $-\pi \leq H \leq \pi$ as an input to the perception network.

The occupancy map $O$ is encoded using four 2D convolutions with batched normalization, adaptive average pooling, and a series of fully connected (FC) layers to produce a single embedding. Each timestep of the dynamic obstacle input $D$ is passed through a similar architecture as the occupancy map encoder but utilizing 1D convolutions instead, resulting in $T$ vectors. These are then passed through an LSTM and an FC layer to produce a single embedding. The LSTM is utilized only to encode dynamic obstacle data for the past $T (=5)$ timesteps and not for obstacle trajectory prediction. Finally, the heading $H$ is processed through an FC layer and concatenated with the occupancy map and dynamic obstacle embedding. This concatenated vector is then passed through an FC layer for the final conditioning embedding $\mathcal{O}$.

\section{Validation and Benchmarking}
The objective of this section is to answer the following questions.
\begin{enumerate}
    \item Does our method generalize to multiple pedestrian densities and environments?
    \item How does our method compare to the current state-of-the-art methods?
    \item Is our method reliant on a global plan?
\end{enumerate}

\subsection{Implementation Details}
\noindent The VQVAE codebook is set to 64 vectors, of size 4 each. The input occupancy map is generated using a 5m maximum observation radius and a 0.1 resolution.
The number of PRIEST iterations is set to 2 (6x lower than the original), with 50 initial samples drawn from the distribution predicted by the PixelCNN. The PRIEST optimization is done using a maximum of 10 dynamic obstacles and 100 static obstacles (from the downsampled point cloud). All inference is done on an NVIDIA 3060 Mobile GPU with 6GB of VRAM but occupies less than 1GB in practice.

\subsection{Simulation Configuration}
\subsubsection{Robot Configuration}
We use the Turtlebot2 Robot equipped with a 2D LIDAR sensor to map the environments before runs and as the primary sensor to navigate the environment during each run. We use the open-source AMCL \cite{amcl} library for localization during trials.

\subsubsection{Environment Configuration}
We naturally test in the Lobby World environment where our pipeline has been trained. The other unseen environments we test include Cumberland, Freiburg, Autolab, and Square World, as described in the DRL-VO paper, and an additional custom Hospital World. All these environments had 35 dynamic pedestrians, and in the Lobby, Cumberland, and Square worlds, we additionally tested with 55 dynamic pedestrians.

\subsection{Qualitative Results}
\noindent Figure~\ref{fig_qualitative} shows the typical trajectories resulting from our pipeline in  Cumberland and Lobby environments. The VQ-VAE+PixelCNN trajectories are shown in grey. As can be seen, these trajectories are very diverse but do not necessarily converge at the goal location. The trajectory distribution resulting after refinement from the PRIEST optimizer is shown in green. As can be seen, the refined trajectories are smoother, directed towards the goal, and show multi-modal behaviour for avoiding collisions. We find that 2 iterations of PRIEST optimization are enough for all the experiments reported in this section.



\subsection{Comparison with DRL-VO \cite{xie2023drl}}

\begin{table*}
\label{Table 1}
\centering
\begin{adjustbox}{width=0.83\linewidth,center}
\begin{tblr}{
  colspec = {Q[140]Q[135]Q[130]Q[85]Q[85]Q[85]Q[85]Q[85]Q[85]Q[85]Q[85]},
  cells = {c},
  row{even} = {Gallery},
  row{1} = {Silver},
  cell{1}{1} = {r=2}{},
  cell{1}{2} = {r=2}{},
  cell{1}{3} = {r=2}{},
  cell{1}{4} = {c=2}{},
  cell{1}{6} = {c=2}{},
  cell{1}{8} = {c=2}{},
  cell{1}{10} = {c=2}{},
  cell{3}{1} = {r=6}{Gallery},
  cell{3}{2} = {r=2}{},
  cell{5}{2} = {r=2}{},
  cell{7}{2} = {r=2}{},
  cell{9}{1} = {r=6}{},
  cell{9}{2} = {r=2}{},
  cell{11}{2} = {r=2}{},
  cell{13}{2} = {r=2}{},
  hline{1,3,9,15} = {-}{1.1pt},
  hline{2} = {4-11}{},
  hline{4,6,8,10,12,14} = {3-11}{Silver},
  hline{5,7,11,13} = {2-11}{},
  vlines,
  vline{1,2,3,4,6,8,10,12} = {-}{1pt},
}
\textbf{Planner Type}                     & \textbf{Environment}  & \textbf{Method}               & \textbf{Success Rate ↑} &               & \textbf{Average Time [s] ↓} &                & \textbf{Average Length [m] ↓} &                & \textbf{Average Speed [m/s] ↑} &               \\
                                 &              &                      & 35             & 55            & 35                 & 55             & 35                   & 55             & 35                    & 55            \\
{w/ Global Planner\\ (Dijkstra)} & Lobby World  & DRL-VO               & 0.65           & 0.55          & 19.06              & 23.68          & \textbf{12.2}        & \textbf{13.5}  & 0.64                  & 0.57          \\
                                 &              & \textbf{CrowdSurfer} & \textbf{0.85}  & \textbf{0.80} & \textbf{17.58}     & \textbf{21.67} & 12.66                & 13.65          & \textbf{0.72}         & \textbf{0.63} \\
                                 & Cumberland   & DRL-VO               & 0.75           & 0.55          & 16.57              & \textbf{18.57} & \textbf{10.11}       & \textbf{11.14} & 0.61                  & 0.60          \\
                                 &              & \textbf{CrowdSurfer} & \textbf{0.90}  & \textbf{0.85} & \textbf{15.16}     & 19.73          & 11.37       & 12.82          & \textbf{0.75}         & \textbf{0.65} \\
                                 & Square World & DRL-VO               & 0.80           & 0.80          & 19.81              & 26.39          & 14.86                & 18.74          & 0.75                  & \textbf{0.71} \\
                                 &              & \textbf{CrowdSurfer} & \textbf{0.90}  & \textbf{0.90} & \textbf{18.5}      & \textbf{24.55} & \textbf{14.43}       & \textbf{16.94} & \textbf{0.78}         & 0.69          \\
w/o Global Planner               & Lobby World  & DRL-VO               & 0.45           & 0.40          & 25.41              & \textbf{26.47} & \textbf{16.01}       & \textbf{14.29} & 0.63                  & 0.54          \\
                                 &              & \textbf{CrowdSurfer} & \textbf{0.75}  & \textbf{0.65} & \textbf{23.44}     & 28.75          & 16.17                & 18.11          & \textbf{0.69}         & \textbf{0.63} \\
                                 & Cumberland   & DRL-VO               & 0.50           & 0.35          & 22.70              & 41.31          & 13.62                & 19.83          & 0.60                  & \textbf{0.48} \\
                                 &              & \textbf{CrowdSurfer} & \textbf{0.75}  & \textbf{0.65} & \textbf{21.22}     & \textbf{26.59} & \textbf{13.37}       & \textbf{17.20} & \textbf{0.63}         & 0.47          \\
                                 & Square World & DRL-VO               & 0.65           & 0.60          & 31.40              & 35.86          & 22.61                & 23.67          & 0.72                  & \textbf{0.66} \\
                                 &              & \textbf{CrowdSurfer} & \textbf{0.80}  & \textbf{0.80} & \textbf{29.61}     & \textbf{32.71} & \textbf{21.91}       & \textbf{21.26} & \textbf{0.74}         & 0.65          
\end{tblr}
\end{adjustbox}

\caption{\footnotesize{Navigation results for three selected testing environments with 35 and 55 dynamic pedestrians. (The averaged values are calculated only for successful scenarios). Our pipeline shows a significant improvement in success rate in the navigational tasks performed in densely crowded scenarios as opposed to DRL-VO.}}
\label{Table1}
\vspace{-0.6cm}
\end{table*}

\noindent In order to measure the efficacy of our method, we compare it against DRL-VO \cite{xie2023drl}, a state-of-the-art dynamic scene navigation algorithm. We provide test results using the Gazebo simulator and the PEDSIM library in 3 distinct environments and different densities of pedestrians. We first consider performance when the global plan is provided to replicate the original DRL-VO testing conditions exactly. We compare using four commonly used metrics in the literature:
\begin{enumerate}
    \item \textbf{Success Rate:} the fraction of trials where the robot reaches the goal without any collisions.
    \item \textbf{Average Time:} average travel time across trials
    \item \textbf{Average Length:} average length of the entire path to goal across trials
    \item \textbf{Average Speed:} average travel speed across trials
\end{enumerate}

\noindent For each environment and pedestrian density, we assign 20 successive goal points. Each goal point is a single trial. A trial is considered successful if the goal is reached without collisions. In case of failure, the current goal is used as the initial position, and trials continue from the next goal. The averaged speed and distance metrics are computed exclusively for successful trials where the agent successfully reached its designated goal.

Table \ref{Table1} shows the previously described metrics for three of the testing environments. Cumberland and Square World were unseen environments for our VQ-VAE+PixelCNN model. As can be seen, we show strong success rates on Lobby World, which was seen during training. However, more importantly, our performance generalizes well to the unseen scenarios. In all environments, our method consistently outperforms DRL-VO with a success rate upwards of 0.8 with both 35 and 55 pedestrians. This is approximately $40 \%$ improvement over DRL-VO.

Additionally, our pipeline is able to traverse the benchmark scenarios with a higher average speed, and a lower average path length in most cases than those observed with DRL-VO. This is indicative of the reactive nature of DRL-VO, as compared to the long-horizon planning afforded by our pipeline. In some cases, as in Cumberland and Square World environments, DRL-VO takes a lower trajectory length by showing more aggressive planning behaviour. However, our pipeline still yields a lower average time in these cases by planning ahead. 

It is important to highlight that the DRL-VO performance, especially the success rate we obtained, is substantially lower than that reported in \cite{xie2023drl}. One reason for this is that we benchmark with substantially longer runs than \cite{xie2023drl}. Specifically, the start-to-goal distance in our benchmarking is almost twice what is used in \cite{xie2023drl}.

Due to its one-shot planning approach and reactive nature, DRL-VO compute times were typically lower at approximately 0.022 seconds. However, our pipeline has a compute time adequately low for real-time functioning of approximately 0.041 seconds despite calculating trajectories over a horizon of 5 seconds with 50 time-steps, as seen in Fig. \ref{fig_graph}.


\vspace{-0.5mm}

\subsection{Adapting to Changing Map}
\noindent In this subsection, we test the adaptivity of our approach and DRL-VO to changes in the static environment. We take the Square World environment with 25 dynamic pedestrians and add additional static obstacles. These obstacles were not seen during the mapping, and thus, the prior-compute plan became infeasible on many runs. Table \ref{Table2} summarizes the key results for 9 trials. As can be seen, the success rate of DRL-VO dropped from 0.80 (Table \ref{Table1}) to 0.44. In contrast, our approach showed a marginal drop to 0.89 from 0.90 (Table \ref{Table1}).

\vspace{-0.3mm}

\subsection{Reliance on Global Plan}
\noindent Table \ref{Table1} (second half) compares the performance of DRL-VO and our approach without guidance from the global plan. More specifically, the global plan is just a straight line connecting the start and the goal. As can be seen, although the success rate reduces for both approaches, the performance falloff is steeper for DRL-VO, dropping approximately 28\% on average. In contrast, our approach is much more robust, dropping approximately 15\% on average.

\vspace{0.1cm}
\subsection{Ablation: Comparison with PRIEST \cite{rastgar2024priest}}

\noindent Our pipeline builds on PRIEST by augmenting it with a learned and environment-conditioned initialization through the VQ-VAE+PixelCNN model. Thus, this sub-section analyzes our learned model's impact over the regular PRIEST pipeline. We consider the Cumberland environment with 35 dynamic agents as PRIEST performed best in this setting. The results are summarized in Table \ref{PRIEST Ablation}. We showcase the statistics for different computation budgets of PRIEST (2, 10, 12 iterations). As can be seen, the environment-conditioned priors provided by VQ-VAE+PixelCNN dramatically improve the success rate along with average travel time and speed.

\vspace{0.3cm}
\vspace*{-\baselineskip}
\begin{table}[h]
\renewcommand{\arraystretch}{1.2}
\centering
\begin{tabular}{|c|c|}
\hline
\textbf{Method} & \textbf{Success Rate $\uparrow$} \\
\hline
DRL-VO & 0.44 \\
\hline
Ours &  \textbf{0.89}\\
\hline
\end{tabular}
\caption{\footnotesize{Navigation Results with continuously changing environments}}
\label{Table2}
\end{table}

\vspace{-0.1cm}
\vspace*{-\baselineskip}
\begin{table}[h]
\renewcommand{\arraystretch}{1.2}
\setlength{\tabcolsep}{1pt} 
\centering
\begin{tabular}{|c|c|c|c|c|}
\hline
\textbf{Method} & \textbf{Success Rate $\uparrow$} & \textbf{Time[s] $\downarrow$} & \textbf{Length[m] $\downarrow$} & \textbf{Speed[m/s] $\uparrow$} \\
\hline
PRIEST-2 & 0.20 & 62.72 & 18.19 & 0.29 \\
\hline
PRIEST-10 & 0.30 & 56.00 & \textbf{16.24} & 0.29\\
\hline
PRIEST-12 & 0.60 & 53.79 & 17.75 & 0.33 \\
\hline
Ours & \textbf{0.90} & \textbf{31.40} & 23.5 & \textbf{0.75} \\
\hline
\end{tabular}
\caption{\footnotesize{Ablation Study with PRIEST planner \cite{rastgar2024priest}. The PRIEST planner is run for different iterations (2, 10, 12)}}
\label{PRIEST Ablation}
\end{table}

\vspace{-0.3cm}
\subsection{Real World Configuration}
We have tested our pipeline in the real world on two different robots:
\begin{itemize}
    \item The Husky A200 mobile robot, which we have equipped with a SLAMTEC RPLIDAR S2 and an Intel® Realsense D455 RGBD Camera as the primary sensors
    \item A custom-made Autonomous Wheelchair equipped with a SLAMTEC RPLIDAR A3 and an Intel® Realsense D455 RGBD Camera as the primary sensors 
\end{itemize}
For the detection of humans in the scene, we use two different methods and play around:
\begin{itemize}
    \item A combination of YOLO \cite{yolov8} in the image frame and lidar for accurate depth information.
    \item Leg Tracker\cite{7139259}: Detects leg-like patterns in lidar scans using clustering.
\end{itemize}
The testing was done at RRC, IIIT Hyderabad, as shown in Figure~\ref{fig:fig_teaser}. We were able to achieve reasonable success rates both indoors and outdoors, albeit lower than the simulation testing. One of the many reasons could be inaccuracies in the human-detection pipeline, which can be improved in the future using more sophisticated methods. While 2 CEM iterations delivered reasonable success rates, increasing the number of iterations to 3 achieved a better success rate in the real world.

\vspace{-0.2cm}
\section{Conclusion}

This paper proposed a novel framework that couples generative prior models learned on expert data with inference-time optimization to show results that are significantly better than prior art in highly populated pedestrian worlds containing as many as 55 moving pedestrians. While the generative priors are efficient in exploring the homotopies and are inherently multimodal, they are not goal-conditioned and are not guaranteed to avoid collisions. The method of inference-time batch projection ensures both these conditions are met with a high degree of accuracy. Further, the efficacy of the proposed framework in the absence of a global plan or in changed environments where the planner is rendered ineffective is a cornerstone of this effort. In the future, we expect to condition the generative prior on both image and range data even as we focus on improving real-time performance with higher fidelity. 

\section*{Acknowledgements}
The author, Laksh Nanwani, thanks IHub-Data, IIIT Hyderabad, for extending their
research fellowship. We also acknowledge IHub-Data for supporting this work via project M2-029. This work was also funded in part by the European Social Fund and Estonian Research Council via project TEM-TA101 and Grant PSG753 from the Estonian Research Council.

\bibliography{references}

\begin{thebibliography}{10}
\providecommand{\url}[1]{#1}
\csname url@samestyle\endcsname
\providecommand{\newblock}{\relax}
\providecommand{\bibinfo}[2]{#2}
\providecommand{\BIBentrySTDinterwordspacing}{\spaceskip=0pt\relax}
\providecommand{\BIBentryALTinterwordstretchfactor}{4}
\providecommand{\BIBentryALTinterwordspacing}{\spaceskip=\fontdimen2\font plus
\BIBentryALTinterwordstretchfactor\fontdimen3\font minus \fontdimen4\font\relax}
\providecommand{\BIBforeignlanguage}[2]{{%
\expandafter\ifx\csname l@#1\endcsname\relax
\typeout{** WARNING: IEEEtran.bst: No hyphenation pattern has been}%
\typeout{** loaded for the language `#1'. Using the pattern for}%
\typeout{** the default language instead.}%
\else
\language=\csname l@#1\endcsname
\fi
#2}}
\providecommand{\BIBdecl}{\relax}
\BIBdecl

\bibitem{rastgar2024priest}
F.~Rastgar, H.~Masnavi, B.~Sharma, A.~Aabloo, J.~Swevers, and A.~K. Singh, ``Priest: Projection guided sampling-based optimization for autonomous navigation,'' \emph{IEEE Robotics and Automation Letters}, 2024.

\bibitem{xie2023drl}
Z.~Xie and P.~Dames, ``Drl-vo: Learning to navigate through crowded dynamic scenes using velocity obstacles,'' \emph{IEEE Transactions on Robotics}, vol.~39, no.~4, pp. 2700--2719, 2023.

\bibitem{valner2022scalable}
R.~Valner, H.~Masnavi, I.~Rybalskii, R.~P{\~o}llu{\"a}{\"a}r, E.~K{\~o}iv, A.~Aabloo, K.~Kruusam{\"a}e, and A.~K. Singh, ``Scalable and heterogenous mobile robot fleet-based task automation in crowded hospital environments—a field test,'' \emph{Frontiers in Robotics and AI}, vol.~9, p. 922835, 2022.

\bibitem{pratkanis2013replacing}
A.~Pratkanis, A.~E. Leeper, and K.~Salisbury, ``Replacing the office intern: An autonomous coffee run with a mobile manipulator,'' in \emph{2013 IEEE International Conference on Robotics and Automation}.\hskip 1em plus 0.5em minus 0.4em\relax IEEE, 2013, pp. 1248--1253.

\bibitem{xiao2022autonomous}
X.~Xiao, Z.~Xu, Z.~Wang, Y.~Song, G.~Warnell, P.~Stone, T.~Zhang, S.~Ravi, G.~Wang, H.~Karnan \emph{et~al.}, ``Autonomous ground navigation in highly constrained spaces: Lessons learned from the barn challenge at icra 2022,'' \emph{arXiv preprint arXiv:2208.10473}, 2022.

\bibitem{bi2018navigation}
J.~Bi, T.~Xiao, Q.~Sun, and C.~Xu, ``Navigation by imitation in a pedestrian-rich environment,'' \emph{arXiv preprint arXiv:1811.00506}, 2018.

\bibitem{chen2019crowd}
C.~Chen, Y.~Liu, S.~Kreiss, and A.~Alahi, ``Crowd-robot interaction: Crowd-aware robot navigation with attention-based deep reinforcement learning,'' in \emph{2019 international conference on robotics and automation (ICRA)}.\hskip 1em plus 0.5em minus 0.4em\relax IEEE, 2019, pp. 6015--6022.

\bibitem{zhou2022robot}
Z.~Zhou, P.~Zhu, Z.~Zeng, J.~Xiao, H.~Lu, and Z.~Zhou, ``Robot navigation in a crowd by integrating deep reinforcement learning and online planning,'' \emph{Applied Intelligence}, vol.~52, no.~13, pp. 15\,600--15\,616, 2022.

\bibitem{chen2020relational}
C.~Chen, S.~Hu, P.~Nikdel, G.~Mori, and M.~Savva, ``Relational graph learning for crowd navigation,'' in \emph{2020 IEEE/RSJ International Conference on Intelligent Robots and Systems (IROS)}.\hskip 1em plus 0.5em minus 0.4em\relax IEEE, 2020, pp. 10\,007--10\,013.

\bibitem{van2017neural}
A.~Van Den~Oord, O.~Vinyals \emph{et~al.}, ``Neural discrete representation learning,'' \emph{Advances in neural information processing systems}, vol.~30, 2017.

\bibitem{van2016conditional}
A.~Van~den Oord, N.~Kalchbrenner, L.~Espeholt, O.~Vinyals, A.~Graves \emph{et~al.}, ``Conditional image generation with pixelcnn decoders,'' \emph{Advances in neural information processing systems}, vol.~29, 2016.

\bibitem{shafabakhsh2013simulation}
G.~Shafabakhsh and M.~Mohammadi, ``Simulation of pedestrian movements using social force model,'' \emph{Journal of Modeling in Engineering}, vol.~11, no.~34, pp. 49--62, 2013.

\bibitem{van2008reciprocal}
J.~Van~den Berg, M.~Lin, and D.~Manocha, ``Reciprocal velocity obstacles for real-time multi-agent navigation,'' in \emph{2008 IEEE international conference on robotics and automation}.\hskip 1em plus 0.5em minus 0.4em\relax Ieee, 2008, pp. 1928--1935.

\bibitem{fox1997dynamic}
D.~Fox, W.~Burgard, and S.~Thrun, ``The dynamic window approach to collision avoidance,'' \emph{IEEE Robotics \& Automation Magazine}, vol.~4, no.~1, pp. 23--33, 1997.

\bibitem{rosmann2015timed}
C.~R{\"o}smann, F.~Hoffmann, and T.~Bertram, ``Timed-elastic-bands for time-optimal point-to-point nonlinear model predictive control,'' in \emph{2015 european control conference (ECC)}.\hskip 1em plus 0.5em minus 0.4em\relax IEEE, 2015, pp. 3352--3357.

\bibitem{williams2017model}
G.~Williams, A.~Aldrich, and E.~A. Theodorou, ``Model predictive path integral control: From theory to parallel computation,'' \emph{Journal of Guidance, Control, and Dynamics}, vol.~40, no.~2, pp. 344--357, 2017.

\bibitem{mohamed2022autonomous}
I.~S. Mohamed, K.~Yin, and L.~Liu, ``Autonomous navigation of agvs in unknown cluttered environments: log-mppi control strategy,'' \emph{IEEE Robotics and Automation Letters}, vol.~7, no.~4, pp. 10\,240--10\,247, 2022.

\bibitem{everett2018motion}
M.~Everett, Y.~F. Chen, and J.~P. How, ``Motion planning among dynamic, decision-making agents with deep reinforcement learning,'' in \emph{2018 IEEE/RSJ International Conference on Intelligent Robots and Systems (IROS)}.\hskip 1em plus 0.5em minus 0.4em\relax IEEE, 2018, pp. 3052--3059.

\bibitem{9197379}
A.~J. Sathyamoorthy, J.~Liang, U.~Patel, T.~Guan, R.~Chandra, and D.~Manocha, ``Densecavoid: Real-time navigation in dense crowds using anticipatory behaviors,'' in \emph{2020 IEEE International Conference on Robotics and Automation (ICRA)}, 2020, pp. 11\,345--11\,352.

\bibitem{9099106}
A.~J. Sathyamoorthy, U.~Patel, T.~Guan, and D.~Manocha, ``Frozone: Freezing-free, pedestrian-friendly navigation in human crowds,'' \emph{IEEE Robotics and Automation Letters}, vol.~5, no.~3, pp. 4352--4359, 2020.

\bibitem{hirose2023sacson}
N.~Hirose, D.~Shah, A.~Sridhar, and S.~Levine, ``Sacson: Scalable autonomous control for social navigation,'' \emph{IEEE Robotics and Automation Letters}, vol.~9, no.~1, pp. 49--56, 2024.

\bibitem{8011466}
X.-T. Truong and T.~D. Ngo, ``Toward socially aware robot navigation in dynamic and crowded environments: A proactive social motion model,'' \emph{IEEE Transactions on Automation Science and Engineering}, vol.~14, no.~4, pp. 1743--1760, 2017.

\bibitem{6698863}
G.~Ferrer, A.~Garrell, and A.~Sanfeliu, ``Social-aware robot navigation in urban environments,'' in \emph{2013 European Conference on Mobile Robots}, 2013, pp. 331--336.

\bibitem{10611710}
A.~H. Raj, Z.~Hu, H.~Karnan, R.~Chandra, A.~Payandeh, L.~Mao, P.~Stone, J.~Biswas, and X.~Xiao, ``Rethinking social robot navigation: Leveraging the best of two worlds,'' in \emph{2024 IEEE International Conference on Robotics and Automation (ICRA)}, 2024, pp. 16\,330--16\,337.

\bibitem{10610025}
P.~Roth, J.~Nubert, F.~Yang, M.~Mittal, and M.~Hutter, ``Viplanner: Visual semantic imperative learning for local navigation,'' in \emph{2024 IEEE International Conference on Robotics and Automation (ICRA)}, 2024, pp. 5243--5249.

\bibitem{idoko2024learning}
S.~Idoko, B.~Sharma, and A.~K. Singh, ``Learning sampling distribution and safety filter for autonomous driving with vq-vae and differentiable optimization,'' \emph{arXiv preprint arXiv:2403.19461}, 2024.

\bibitem{oord2017neural}
A.~v.~d. Oord, O.~Vinyals, and K.~Kavukcuoglu, ``Neural discrete representation learning,'' \emph{arXiv preprint arXiv:1711.00937}, 2017.

\bibitem{wen2018constrained}
M.~Wen and U.~Topcu, ``Constrained cross-entropy method for safe reinforcement learning,'' \emph{Advances in Neural Information Processing Systems}, vol.~31, 2018.

\bibitem{gloor-2016}
\BIBentryALTinterwordspacing
C.~Gloor, ``{PEDSIM: Pedestrian crowd simulation},'' 2016. [Online]. Available: \url{http://pedsim.silmaril.org}
\BIBentrySTDinterwordspacing

\bibitem{koenig2004design}
N.~Koenig and A.~Howard, ``Design and use paradigms for gazebo, an open-source multi-robot simulator,'' in \emph{2004 IEEE/RSJ international conference on intelligent robots and systems (IROS)(IEEE Cat. No. 04CH37566)}, vol.~3.\hskip 1em plus 0.5em minus 0.4em\relax Ieee, 2004, pp. 2149--2154.

\bibitem{amcl}
\BIBentryALTinterwordspacing
B.~P. Gerkey, ``Amcl - ros wiki.'' [Online]. Available: \url{https://wiki.ros.org/amcl}
\BIBentrySTDinterwordspacing

\bibitem{yolov8}
G.~Jocher, A.~Chaurasia, and J.~Qiu, ``Yolo by ultralytics,'' \url{https://github.com/ultralytics/ultralytics}, 2023.

\bibitem{7139259}
A.~Leigh, J.~Pineau, N.~Olmedo, and H.~Zhang, ``Person tracking and following with 2d laser scanners,'' in \emph{2015 IEEE International Conference on Robotics and Automation (ICRA)}, 2015, pp. 726--733.

\end{thebibliography}
\bibliographystyle{IEEEtran}

\end{document}